\documentclass[journal]{IEEEtran}
\usepackage{graphicx}
\usepackage{hyperref}

\begin{document}
\title{Evaluation of ChatGPT for NLP-based Mental Health Applications}

\author{Bishal Lamichhane,\\ lamichhane.bishal@gmail.com}% <-this % stops a space

\maketitle

\begin{abstract}
Large language models (LLM) have been successful in several natural language understanding tasks and could be relevant for natural language processing (NLP)-based mental health application research. In this work, we report the performance of LLM-based ChatGPT (with gpt-3.5-turbo backend) in three text-based mental health classification tasks: stress detection (2-class classification), depression detection (2-class classification), and suicidality detection (5-class classification). We obtained annotated social media posts for the three classification tasks from public datasets. Then ChatGPT API classified the social media posts with an input prompt for classification. We obtained F1 scores of 0.73, 0.86, and 0.37 for stress detection, depression detection, and suicidality detection, respectively. A baseline model that always predicted the dominant class resulted in F1 scores of 0.35, 0.60, and 0.19. The zero-shot classification accuracy obtained with ChatGPT indicates a potential use of language models for mental health classification tasks.  
\end{abstract}

% Note that keywords are not normally used for peerreview papers.

\IEEEpeerreviewmaketitle

\section{Introduction}
Mental health illnesses are widely prevalent. More than one billion people worldwide are estimated to be suffering from mental illnesses~\cite{world2022world}. In the US, one in five individuals has mental health issues every year~\cite{abuse2020key}. Proper management of mental health with timely diagnosis, intervention, and monitoring requires \textit{measuring} mental health. Since most mental health illnesses have bio-psycho-social origins, research on measuring mental health has considered obtaining bio-behavioral signals to detect the mental health state of the individual. Social media posts could be a source to detect the mental health state of an individual. One's mental health state could be reflected in the sentiments or linguistic contents of their social media posts. Accordingly, previous works have demonstrated mental health state detection based on the analysis of users' social media posts~\cite{gkotsis2017characterisation,chancellor2020methods,kim2020deep}.  

Mental health state detection from texts such as social media posts requires natural language processing (NLP) to infer the sentiment and content shared in the text. Recently, large language models (LLM) have shown unprecedented success in natural language understanding. LLMs are neural networks based on the transformer architecture~\cite{vaswani2017attention} trained in a self-supervised setting with a large text corpus, followed by fine-tuning in supervised and reinforcement learning settings. ChatGPT is a popular LLM-based chat application/assistant that has been receiving a lot of public attention. ChatGPT as a conversation agent demonstrates its capability to understand the text and respond accordingly. Though several limitations of ChatGPT and LLMs, in general, are acknowledged, these models have shown to be good generalized models across different applications. The applications of the GPT-4 model, the latest LLM model from OpenAI, in several areas such as programming and mathematics were demonstrated by the authors in ~\cite{bubeck2023sparks}. Similarly, the authors in ~\cite{nori2023capabilities} demonstrated GPT-4's capability in the medical domain for answering USMLE (United States Medical Licensing Examination) and MultMedQA dataset~\cite{singhal2022large} questions.  

One application area for LLM could be in mental health, given the model's capability for language understanding. Different application scenarios could arise for LLM's use in a mental health context. As a user interacts more with LLMs, sometimes even sharing their thoughts and concerns, the posts shared with LLMs could provide a better view of the person's mental health compared to social media posts. The LLMs could adapt to respond based on the person's mental health state or even provide interventions, e.g., by suggesting to seek medical care. LLM's language understanding could also be used as a backend in many front-end mental health applications. Much effort today is focused on building custom machine-learning models for mental health detection tasks based on texts and other modalities. As the LLMs get more intelligent, it is not inconceivable that these LLMs become a defacto backend for all language understanding tasks such as in mental health applications 

Given the current capability of ChatGPT as one of the most popular LLM-based chat applications, in this work, we evaluated ChatGPT's zero-shot classification performance in three mental health application tasks: stress detection, depression detection, and suicidality detection tasks based on user's social media posts. We used publicly available labeled datasets for our evaluations.

\section{Method}

\subsection{Dataset}

\subsubsection{Stress Detection Dataset}
We evaluated ChatGPT's ability to detect stressed states using the labeled stress detection dataset available at \cite{stress_github}. The dataset was inspired by the work in ~\cite{turcan2019dreaddit} and consists of users' social media posts in different Reddit group categories. A train and test set of 2838 posts and 715 posts, respectively, are available in the dataset. Each post has an associated label of stress or non-stress. We evaluated the ChatGPT model on the test set posts only.

An example post in the stress class is:\\ \textit{"By the fourth infusion, I was able to sleep through the night, and only got a mild feeling of anxiety when talking or thinking about it. Today, while leaving my psychologist's office, I saw someone get hit by a car. It was not pretty. He died. Right as I'm starting to move past one, another happens."} \\

Similarly, an example post in the non-stress class is:\\ \textit{"Best friend knows I have anxiety and I am always asking her if she's mad at me. Well, yesterday she asked if I wanted to go to the beach next weekend, I said yes, and then today I told her I couldn't because I was supposed to watch my sister's kids. She wants to know why I don't take them with us. It's a two hour drive, and I really just don't want to go. But then she says that I always say no when she asks me to go somewhere with her."}\\

\subsubsection{Depression Detection Dataset}

For the depression detection task, we used the dataset available at \cite{depression_github}. The dataset consists of user posts in different Reddit groups (subreddits) and web blogs. We used the user posts in depression-related Reddit groups as the data for the \textit{depression} class. The Reddit posts from the \textit{non-depression} category in the dataset~\cite{depression_github} were taken as the negative class. The number of posts in the depression class is 1293 and that in the non-depression class is 548.\\

\subsubsection{Suicidality Detection Dataset}

A suicidality detection dataset based on social media posts is available in ~\cite{suic_github}. We used the subset of 500 users in the dataset obtained from~\cite{gaur2019knowledge}. Experts labeled posts from these users as belonging to one of the five classes: {Supportive, Suicidal Ideation, Suicidal Behavior, Suicidal Attempt, and Suicidal Indicator}.

\subsection{ChatGPT Evaluation}

For each detection task, we provided the input text from the user post to the OpenAI ChatGPT API with the following prompt.\\

\textit{"Which of the classes can the following post be assigned to? Only reply with answers from one of the classes: CLASS\_NAMES.}.\\

For example, for the stress/non-stress detection task, the prompt would be:

\textit{"Which of the classes can the following post be assigned to? Only reply with answers from one of the classes: stressed, non-stressed.}.\\

We used the python \textit{openai} package to invoke the ChatComplete API of ChatGPT. The gpt-3.5-turbo backend was used; we still do not have access to the GPT-4 backend. We compared ChatGPT's detection from a single API call to the annotations in the dataset and computed the F1 score of detection. In a multi-class setting, we compute the weighted F1 score. We also evaluated the confusion matrix of the predictions. 

\section{Results}

The F1 score obtained for the three detection tasks is provided in Table~\ref{tab:clf_table}. We obtained an F1 score of 0.72 for the stress detection task. In this 2-class classification task, the baseline model that predicted the dominant class (stress) resulted in an F1 score of 0.35. For the depression detection task, we obtained an F1 score of 0.86 which is much higher than the baseline F1 score of 0.60 obtained by always predicting the dominant class (depression). Finally, we obtained a low F1 score of 0.37 for the 5-class suicidality classification task. The F1 score was still higher than the baseline F1 score of 0.19 obtained by always predicting the dominant class (Suicidal Ideation).

\begin{table}[!htb]
    \caption{Classification performance obtained in the three text-based mental health applications using ChatGPT.}
    \label{tab:clf_table}

    \centering
    \begin{tabular}{c|c|c}
    \hline
    \textbf{Dataset} & \textbf{F1 score} & \textbf{Balanced Accuracy} \\
    \hline
    Stress Detection &  0.73 & 0.73  \\
    \hline
    Depression Detection & 0.86 & 0.85 \\
    \hline
    Suicidality Detection & 0.37 & 0.33  \\
    \hline
    \end{tabular}
\end{table}

\begin{figure}
    \centering
    \includegraphics[width=\columnwidth]{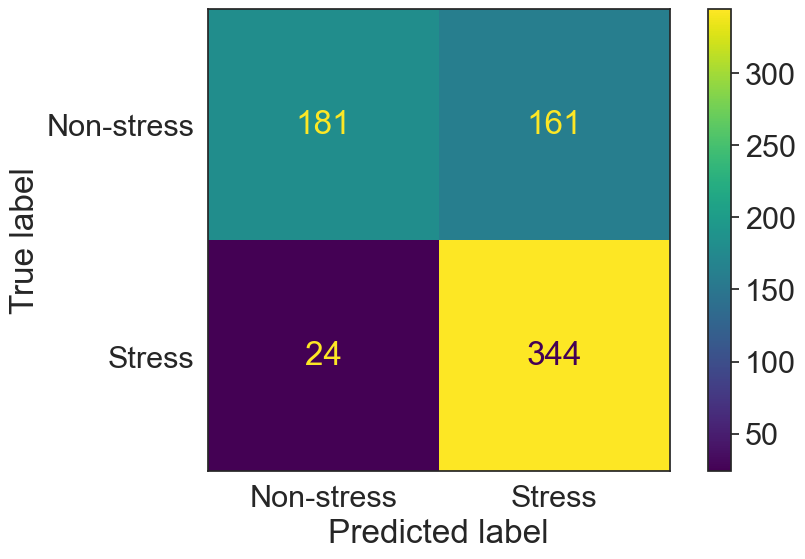}
    \caption{Confusion matrix for the prediction from ChatGPT on the stress detection task.}
    \label{fig:confumat_stress}
\end{figure}

\begin{figure}
    \centering
    \includegraphics[width=\columnwidth]{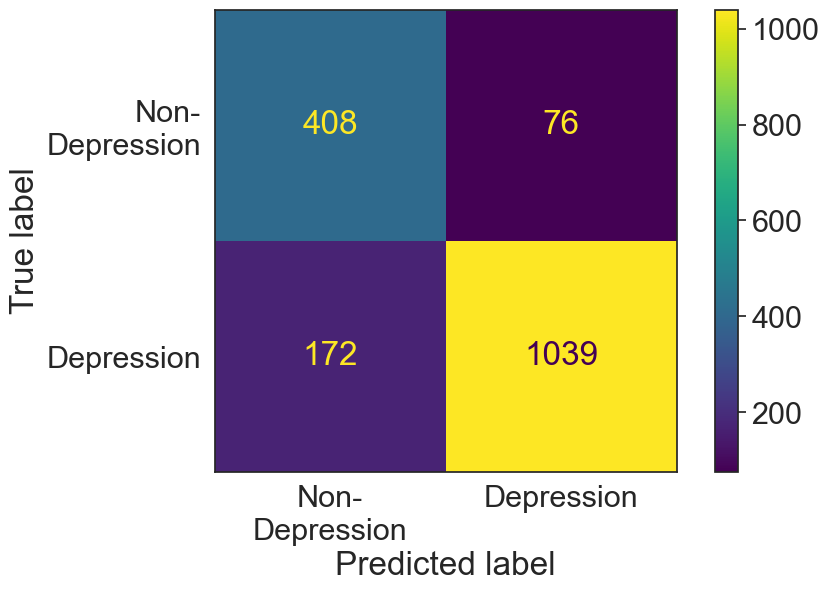}
    \caption{Confusion matrix for the prediction from ChatGPT on the depression detection task.}
    \label{fig:confumat_depr}
\end{figure}

\begin{figure}
    \centering
    \includegraphics[width=\columnwidth]{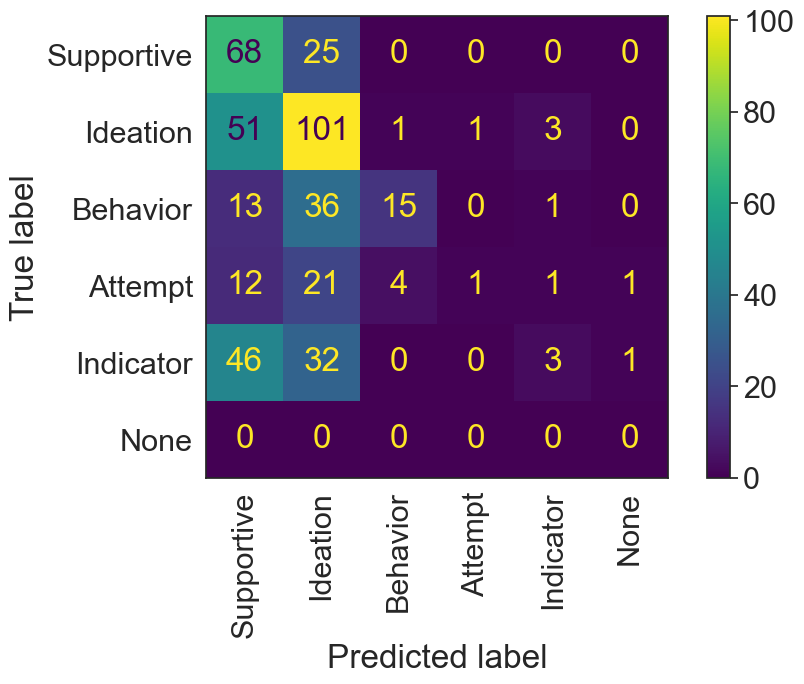}
    \caption{Confusion matrix for the prediction from ChatGPT on the 5-class suicidality detection task. When ChatGPT could not assign any of the five classes to the input text, we labeled it as belonging to the None class.}
    \label{fig:confumat_suic}
\end{figure}

The confusion matrix for each of the three classification tasks is shown in Figure~\ref{fig:confumat_stress},\ref{fig:confumat_depr}, and \ref{fig:confumat_suic} respectively. Some posts in the dataset could not be evaluated due to errors in reading these posts from the dataset or invoking the API with the given posts.

\section{Discussion}

LLMs show a good language understanding capability and could be useful in NLP-based mental health applications. One such application is mental health state detection based on users' social media posts. We evaluated the zero-shot classification accuracy obtained from ChatGPT, a popular LLM-based application, across three mental health detection tasks. We found that ChatGPT provides a good classification performance compared to the baseline model, indicating that ChatGPT can infer a user's mental health state to an extent based on their social media posts. Since the ChatGPT is a general-purpose language model, it could be further fine-tuned/adapted for a specific mental health application or a range of related applications.  

The classification performance obtained from ChatGPT for the stress detection and depression detection task was quite promising (Table~\ref{tab:clf_table}). For the stress detection task, the authors in ~\cite{stress_github} trained three machine learning models using the in-domain training set (labeled user posts similar to the test dataset). The BERT model~\cite{devlin2018bert} fine-tuned for the stress detection task resulted in an F1 score of 0.81, higher than the F1 score obtained with a random forest and logistic regression model trained using the TF-IDF (term frequency-inverse document frequency) and Word2Vec features. The F1 score of the fine-tuned BERT model reported in~\cite{stress_github} is higher than the F1 score of 0.73 obtained from the ChatGPT model in our evaluation (Table~\ref{tab:clf_table}). However, we evaluated ChatGPT in a zero-shot classification setting. Thus, the F1 score obtained with ChatGPT could be considered encouraging. Fine-tuning the underlying LLM model for the stress detection task could further improve the classification. For the depression detection task, the authors in ~\cite{depression_github} did not report any model performance. The authors in ~\cite{tadesse2019detection} reported an F1 score of 0.93 on the depression detection dataset using bigrams, LIWC (Linguistic Inquiry and Word Count dictionary), and LDA (Latent Dirichlet Allocation)-based features in a neural network model. With zero-shot classification from ChatGPT already yielding a high F1 score of 0.86, fine-tuning can further improve performance to match or exceed the performance from \cite{tadesse2019detection}. 

The lowest F1 score among our tasks was obtained for the suicidality detection task. Suicidality detection was a 5-class classification problem which might have caused a high classification error due to the higher inter-class confusion. From the confusion matrix (Figure~\ref{fig:confumat_suic}), we observed that inputs are mostly predicted to be in the suicidal ideation category. Compared to the binary stress and depression detection tasks, the class boundaries between the closely related classes of suicidality (ideation, behavior, attempt, and indicator) might not be as distinct. The authors in ~\cite{gaur2019knowledge} reported an F1 score of 0.65 on the suicidality detection dataset using word embeddings as input to a CNN model. The authors evaluated their model with 5-fold cross-validation. Thus, the model was trained with the in-domain data from four of the five folds. The confusion matrix reported by the authors in ~\cite{gaur2019knowledge} also indicates that suicidal ideation is the most commonly predicted category, similar to our results in Figure ~\ref{fig:confumat_suic}. The high confusion between the suicidality classes obtained in our evaluations and those in ~\cite{gaur2019knowledge} indicates that the class boundary is highly likely to be overlapped. 

Our evaluations have several limitations. First, we evaluated ChatGPT with GPT-3.5-turbo backend. Recently, a much-improved backend of GPT-4 was released. We plan to evaluate the performance of GPT-4 for the mental health application tasks in future work. Second, we only explored a limited prompt setting and considered the first response from the ChatGPT as its prediction. There might be other prompts that can provide better classification. The variations in responses over multiple calls and with different prompts could also provide auxiliary information helpful for the classification tasks. We will explore this in future work. Third, we evaluated detection tasks in relatively small datasets. Future work should consider evaluations using larger text corpora. Finally, the evaluations are highly dependent on the annotations. We used the annotation provided with the stress and depression detection dataset. These two datasets, unlike the suicidality detection dataset, were likely not annotated by multiple experts. The datasets could be re-annotated with experts to understand if the LLM model's confusion aligns with the disagreement between the annotators.  

\bibliographystyle{IEEEtran}
\bibliography{IEEEabrv,main}

\end{document}